\PassOptionsToPackage{numbers,sort&compress}{natbib}
\documentclass{article}

    \usepackage[final]{neurips_2019}

\usepackage[utf8]{inputenc} 
\usepackage[T1]{fontenc}    
\usepackage{hyperref}       
\usepackage{url}            
\usepackage{booktabs}       
\usepackage{amsfonts}       
\usepackage{nicefrac}       
\usepackage{microtype}      

\usepackage{graphicx}
\usepackage{amssymb}
\usepackage{amsmath}
\usepackage{subcaption}
\usepackage{caption}
\usepackage{wrapfig}

\title{Saliency Methods for Explaining Adversarial Attacks}

\author{
Jindong Gu \\
The University of Munich\\
Siemens AG, Corporate Technology\\
\texttt{jindong.gu@siemens.com} \\
\And
Volker Tresp \\
The University of Munich\\
Siemens AG, Corporate Technology\\
\texttt{volker.tresp@siemens.com} \\
}

\begin{document}

\maketitle

\begin{abstract}
The classification decisions of neural networks can be misled by small imperceptible perturbations. This work aims to explain the misled classifications using saliency methods. The idea behind saliency methods is to explain the classification decisions of neural networks by creating so-called saliency maps. Unfortunately, a number of recent publications have shown that many of the proposed saliency methods do not provide insightful explanations. A prominent example is Guided Backpropagation (GuidedBP), which simply performs (partial) image recovery. However, our numerical analysis shows the saliency maps created by GuidedBP do indeed contain class-discriminative information. We propose a simple and efficient way to enhance the saliency maps. The proposed enhanced GuidedBP shows the state-of-the-art performance to explain adversary classifications.
\end{abstract}

\section{Introduction}
\vspace{-0.2cm}
The explanations produced by saliency methods reveal the relationship between inputs and outputs of the underlying model. In image classifications, the explanations are generally visualized as saliency maps. A saliency map (SM) is created using the three components: an input $\pmb{x} \in \mathbb{R}^d$, a model $M$, corresponding to a function $f_x(\cdot)$, and an output class $y_m$. Formally, a saliency map $\pmb{s}^m \in \mathbb{R}^d$ for the classification of the $m$-th class can be defined as
\begin{equation}
\pmb{s}^m = g(\pmb{x}, M, y_m)
\label{equ:sal}
\end{equation}
where $\pmb{s}^m$ has the same dimensions as the input $\pmb{x}$  and $g(\cdot)$ is a function corresponding to a saliency method.
The value of an element $s^m_i$ in $\pmb{s}^{m}$ specifies the relevance of the input feature $x_i$ to the $m$-th class. Here, $m$ could neither denote the ground-truth class nor the class, predicted to be most likely.

In recent years, a large number of saliency methods have been proposed \cite{Simonyan2013DeepIC,zeiler2014visualizing,springenberg2014striving,bach2015pixel,ribeiro2016should,sundararajan2017axiomatic,smilkov2017smoothgrad,shrikumar2017learning,selvaraju2017grad,ancona2017unified,zintgraf2017visualizing,dabkowski2017real,Fong2017InterpretableEO,Gu2018UnderstandingID}.
Notably, \cite{Mahendran2016SalientDN,adebayo2018sanity} show that SMs created by Guided Backpropagation (GuidedBP \cite{springenberg2014striving}) are neither class-discriminative nor sensitive to model parameters. \cite{nie2018theoretical} proves that GuidedBP is essentially doing (partial) image recovery, which is unrelated to the network decisions. In contrast to their conclusions, our numerical analysis shows that the SMs created by GuidedBP do contain class-relevant decisions.

Most of the existing saliency methods in Equation \ref{equ:sal} only consider the SM of the ground-truth class and ignore SMs for the other classes. \cite{alvarez2018robustness} argues that meaningful explanations should be robust to small local perturbations of the input. However, the small perturbation can lead to the misclassification of neural networks \cite{Szegedy2014IntriguingPO,Goodfellow2015ExplainingAH}. After perturbation, we would not expect that the explanations always stay unchanged since the neural networks might make totally different classification decisions. Hence, we propose that saliency methods should be discriminative to adversary perturbation.

Our contributions can be summarised as follows: 1) We identify class-discriminative information in SMs created by GuidedBP and propose a simple and efficient way to enhance the created SMs; 2) We explain classifications of adversary images with the proposed enhanced Guided Backpropagation and the existing saliency methods. The explanations created by these saliency methods are evaluated via qualitative and quantitative experiments.

\section{Enhanced Guided Backpropagation}
\vspace{-0.2cm}
\label{sec:app}
Similar to raw gradient backpropagation, GuidedBP \cite{springenberg2014striving} propagates gradients back to inputs and takes the received gradients as their saliency values. The two methods differ only in handling ReLU layers. In GuidedBP, $G^l = G^{l+1} * \textbf{1}_{G^{l+1} > 0\ and\ X^{l} > 0}$ where $G^{l}$ is the gradients of the $l$-th layer and the $X^{l}$ are the activations before RuLU layer, and $\textbf{1}$ is the indicator function. Since the indicator function filters out parts of the gradients, the gradients received by some input features can be zeros, which is called filtering effect (FE). The filtering effect of an SM is formally defined as $\pmb{s}^m * \textbf{1}_{\pmb{s}^m>0}$.

\cite{alvarez2018robustness} provides a theoretical analysis of GuidedBP. They show that the created SMs of different classes have similar filtering effects, which means that GuidedBP is not class-discriminative. In the following, we show the SMs created by GuiedBP do contain class-discriminative information and propose a simple way to enhance the discriminative information in the corresponding saliency maps.

\begin{figure}
\centering
\begin{minipage}{.49\textwidth}
  \begin{subfigure}[b]{0.45\textwidth}
    \centering
        \includegraphics[scale=0.205]{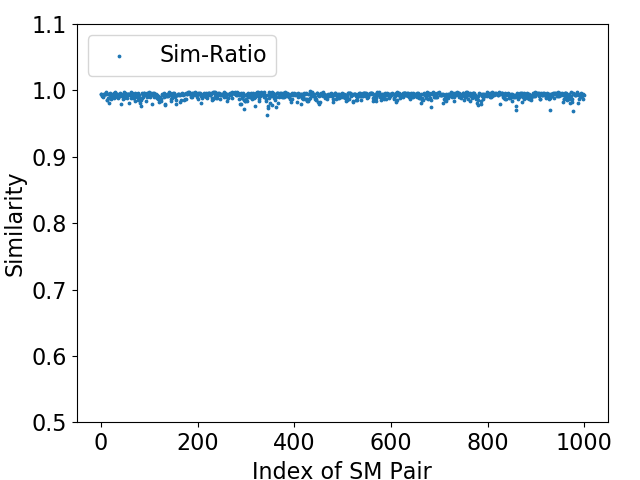}
        \caption{Similarity of FE}
        \label{fig:sim_fe}
    \end{subfigure}
    \hspace{0.05cm}
    \begin{subfigure}[b]{0.52\textwidth}
    \centering
        \includegraphics[scale=0.205]{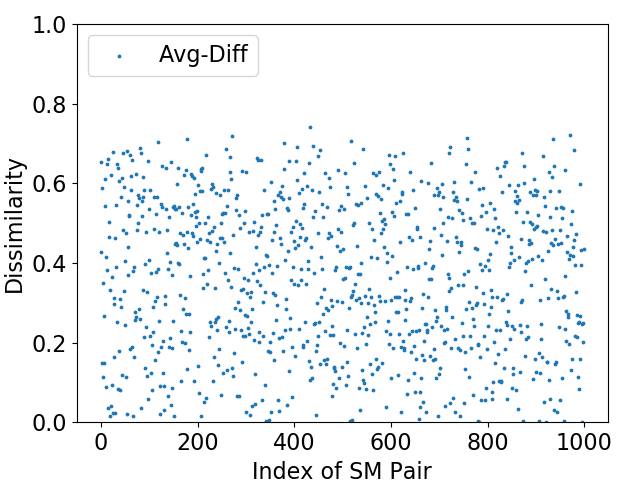}
        \caption{Difference of SM Values}
        \label{fig:avg}
    \end{subfigure}
    \caption{The relationship between two SMs in each SM pair: a) The Sim-Ratio between two binarized SMs describes 
    the similarity of Filtering Effects of them. b) The Avg-Diff between two unnormalized SMs are computed to describe the difference of their saliency values.}
    \label{fig:cluster_vectors}
\end{minipage}%
\hspace{0.15cm}
\begin{minipage}{0.49\textwidth}
  \centering
   \begin{subfigure}{0.48\textwidth}
    \centering
        \includegraphics[scale=0.175]{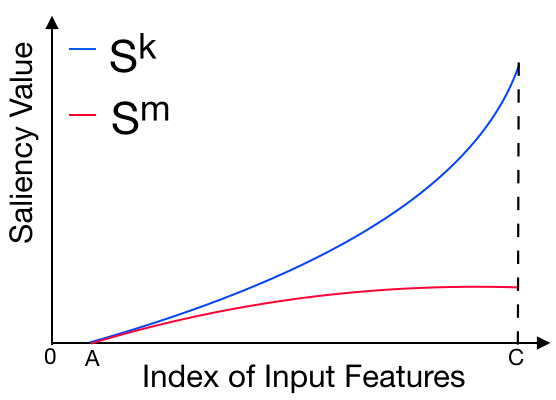}
        \caption{SMs before Norm.}
        \label{fig:before_norm}
    \end{subfigure}
    \hspace{0.05cm}
    \begin{subfigure}{0.48\textwidth}
    \centering
        \includegraphics[scale=0.175]{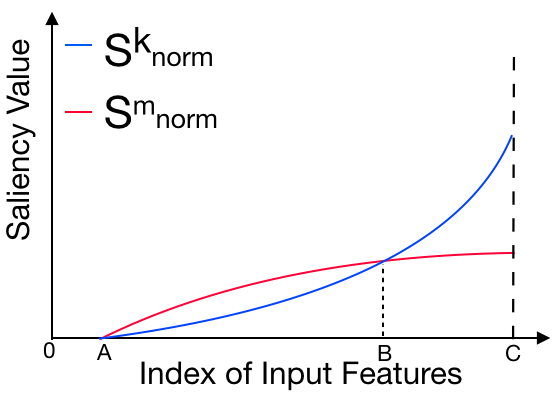}
        \caption{SMs after Norm.}
        \label{fig:after_norm}
    \end{subfigure}
    \caption{This toy example illustrates how the proposed method works to enhance the discriminativity of SMs. In plot b, the indexes located between A and B correspond to the input features relevant to the $m$-th class, and the ones between B and C are the features relevant to the $k$-th class.}
    \label{fig:norm}
\end{minipage}
\end{figure}

\subsection{Identifying Discriminative Information}
\label{sec:dis_info}
$\pmb{s}^k$ and $\pmb{s}^m$ are the two saliency maps created by GuidedBP for the $k$-th output class and the $m$-th output class. They have similar filtering effects, as theoretically analyzed in \cite{alvarez2018robustness}. The difference between them can only be their saliency values, if existing. However, in all published work, SMs are visualized by normalizing saliency values in an SM and mapping them to a color map $[0, 255]$. The possible difference between their saliency values is hidden by the normalization.

In our experiments, we take a pre-trained VGG16 \cite{simonyan2014very} model and fine-tune it on the PASCAL VOC2012 \cite{everingham2010pascal} dataset. Each image in the dataset may have many objects belonging to more than one class. We select images with multiple labels from the validation dataset. For each image, we produce $n$ SMs for $n$ ground-truth classes and choose any two of $n$ SMs to form an SM pair ($\pmb{s}^k$ and $\pmb{s}^m$), i.e., we have $C^2_n$ SM pairs.

For each pair, we compute the similarity between two binarized SMs, which is defined as the ratio between the number of pixels with the same value and the number of all pixels. The scores of all SM pairs in the validation images are shown in Figure \ref{fig:sim_fe}. All the scores are close to 1, which means the SMs of different classes have almost the same filtering effect.

Without normalizing values of SMs, we compute their averages. The difference between the two SMs is defined as Avg-Diff {\small $= \frac{|Avg1 - Avg2|}{\max(Avg1, Avg2)}$}. The scores vary from 0 to 0.8 in Figure \ref{fig:avg}. In summary, given a classification, the two SMs $\pmb{s}^k$ and $\pmb{s}^m$ differ in saliency values instead of filtering effect.

\subsection{Enhancing Discriminative Information of Saliency Maps}
\label{sec:norm}
In this section, we propose a simple and efficient way to extract information about the difference between the two SMs $\pmb{s}^k$ and $\pmb{s}^m$. We argue that the relatively larger saliency values in SMs correspond to the input features that support a specific class. We extract such class-relevant information by normalizing the two SMs and subtracting one by another, which is visualized in Figure \ref{fig:norm}. Figure \ref{fig:before_norm} shows the saliency values of two SMs where input features are ordered by the saliency values of an SM $\pmb{s}^k$. The two SMs have zeros in the interval [0, A] since both have the same filtering effect. The difference between the two SMs is their saliency values in the interval (A, C]. Figure \ref{fig:after_norm} shows the normalized saliency values where the input features of (A, B] are relevant to the $m$-th class, and the ones in (B, C] are relevant to the $k$-th class.

In classifications of real-world images, the obtained discriminative pixels {$max(0, (\pmb{s}_{norm}^k - \pmb{s}_{norm}^m))$} for $k$-th class strongly depend on how the SMs are normalized.  A trivial normalization is to divide the SM by its maximum. However, the maximal value of the SMs (i.e., the maximal local gradient value in vanilla Gradient approach) are noisy and often outliers \cite{Szegedy2014IntriguingPO,smilkov2017smoothgrad}.

An alternative is the energy-based normalization. The individual SMs are normalized by the sum of its saliency values $|\pmb{s}^k|$ (i.e., the energy of the SMs). The SMs {$\pmb{s}^k = (\pmb{s}^k_r, \pmb{s}^k_g, \pmb{s}^k_b)$} and { $\pmb{s}^m = (\pmb{s}^m_r, \pmb{s}^m_g, \pmb{s}^m_b)$} are composed of three channels. The discriminative pixels for the $k$-th class on the R channel are { $Dis^k_r  = max(0, (\frac{\pmb{s}^k_r}{|\pmb{s}^k|} - \frac{\pmb{s}^m_r}{|\pmb{s}^m|})) = max(0, (\frac{\pmb{s}^k_r}{|\pmb{s}^k_r|+|\pmb{s}^k_g|+|\pmb{s}^k_b|} - \frac{\pmb{s}^m_r}{|\pmb{s}^m_r| + |\pmb{s}^m_g| +|\pmb{s}^m_b|}))$}.

Neural networks have different sensitivity to different feature maps and input channels. In a classification, the sensitivity of channels could be different for different output classes. E.g., in case of { $\frac{|\pmb{s}^k_r|}{|\pmb{s}^k_r|+|\pmb{s}^k_g|+|\pmb{s}^k_b|} \ll \frac{|\pmb{s}^m_r|}{|\pmb{s}^m_r|+|\pmb{s}^m_g|+|\pmb{s}^k_b|}$}, the discriminative region { $Dis^k_r = 0$}, and we lose all the information on the red channel. On the contrary case, we might keep too much detail information without highlighting discriminative features. On other channels, we could similarly lose all the information or keep too much non-discriminative information.

We propose the channel-wise energy-based normalization to circumvent the problem. We consider three channels separately. The discriminative pixels of R channel is {$Dis^k_r = max(0, (\frac{\pmb{s}^k_r}{|\pmb{s}^k_r|} - \frac{\pmb{s}^m_r}{|\pmb{s}^m_r|}))$}. Similarly, the discriminative information of each channel is accurately identified. The generalization of the proposed enhancing method to other saliency methods will also be discussed in Section \ref{sec:disc}.

\begin{figure*}
\begin{minipage}{.63\textwidth}
    \centering
    \includegraphics[width=9cm, height=3.5cm]{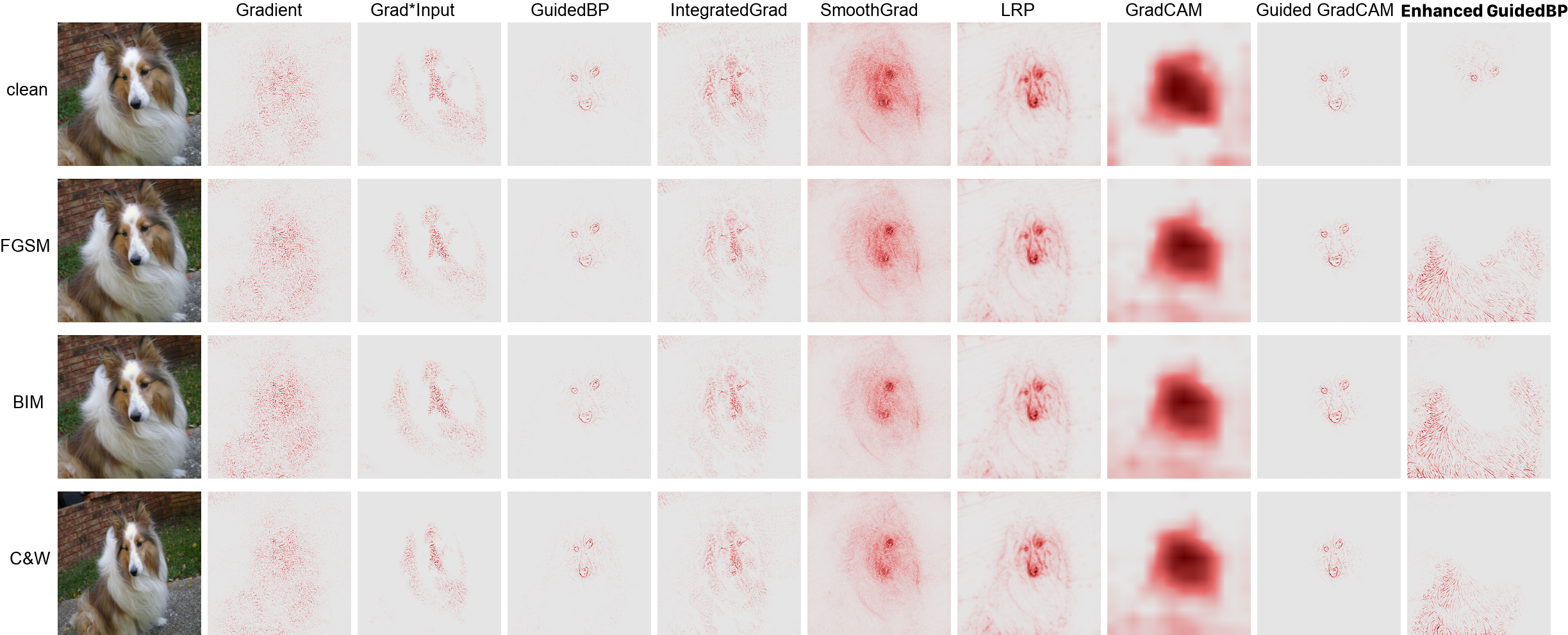}
\caption{This figure shows SMs of clean image and adversary ones. The first column lists the original image and its adversary ones. Our enhanced GuidedBP reacts the adversary attacks strongly, while all other the SMs produce similar SMs.}
    \label{fig:advdis}
   \end{minipage} \hspace{0.2cm}
\begin{minipage}{.35\textwidth}
    \centering
    \includegraphics[width=5cm, height=2.7cm]{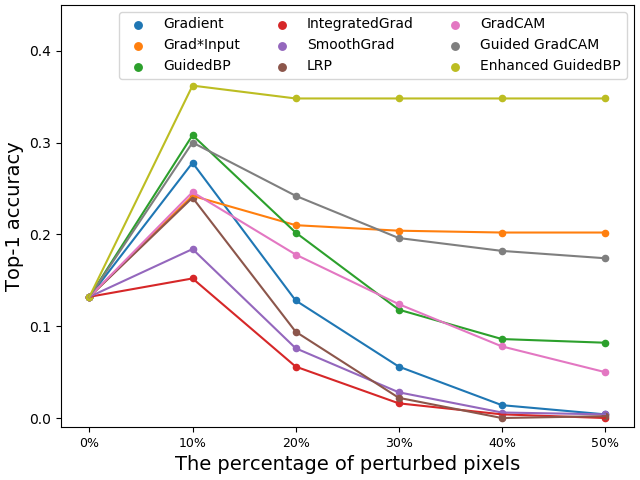}
   \caption{Following the rank of saliency values of a SM, a certain percentage of pixels of the adversary image are perturbed. The classification accuracy on the perturbed adversary images are shown.}
   \label{fig:eval_advdis}
  \end{minipage}
\end{figure*}

\section{Explaining Classifications of Adversary Images}
\vspace{-0.3cm}
Inputs with imperceptible perturbation can fool the well-trained neural networks. The Fast Gradient Sign Method (FGSM) \cite{Szegedy2014IntriguingPO} perturbs an image to increase the loss of classifier on the resulting image. The Basic Iterative Method (BIM) \cite{kurakin2016adversarial} extends FGSM by taking multiple small steps instead of one big step. Another superior attack method is the Carlini and Wagner attack (C\&W) \cite{carlini2017towards}. In the wake of defensive distillation, they create the quasi-imperceptible perturbations by restricting their $l_0, l_2$ and $l_{\inf}$-norms. The $l_2$-norm is used across this paper.

For ImageNet validation images, we create adversary images using the three described attack methods on pre-trained VGG16. The SMs of clean images and adversary images are shown in Figure \ref{fig:advdis}. For all the saliency methods except for our enhanced GuidedBP, the SMs created for predicted classes of the clean image and its adversary versions are visually the same. One might argue that it is an advantage of the saliency methods: they can still identify the object in the image even when attacked. However, we argue that saliency methods should reflect the different decisions of deep neural networks. In other words, they should produce different SMs for clean images and adversary ones.

Since the existing saliency methods always create similar SMs for a clean image and its adversary versions, they cannot be applied to explain classifications misled by adversary perturbations. Our enhanced GuidedBP can identify the relevant evidence of the decisions. For the classification of the original input (e.g., sheepland dog), the created SM shows that the VGG16 focuses on the important visual feature of the target object (i.e., the head), while it focuses on class-irrelevant features (background and body parts) when explaining the classifications of adversary inputs.

The saliency methods can identify the input features that contribute to the classification decision. We can apply saliency methods on misled classifications of adversary samples. If we perturb the pixels relevant to the misclassification according to the created SMs, the attack effectiveness will be decreased. The performance of the model on the perturbated samples can be recovered to some extent. Figure \ref{fig:eval_advdis} shows the performance of the model on the adversary samples (C\&W attack) when they are perturbed according to the SMs. We can observe that the perturbation with SMs of our enhanced GuidedBP can recovery the score better. Instead of claiming that the SM-based perturbation is an effective defense method, we aim to show that SMs created by enhanced GuidedBP can better identify the pixels relevant to classifications. When too many images pixels are perturbed, the visual features of true target objects are lost, which can also lead to low performance of the model.

\begin{wrapfigure}{r}{0.46\textwidth}
\centering
 \includegraphics[width=6cm, height=4cm]{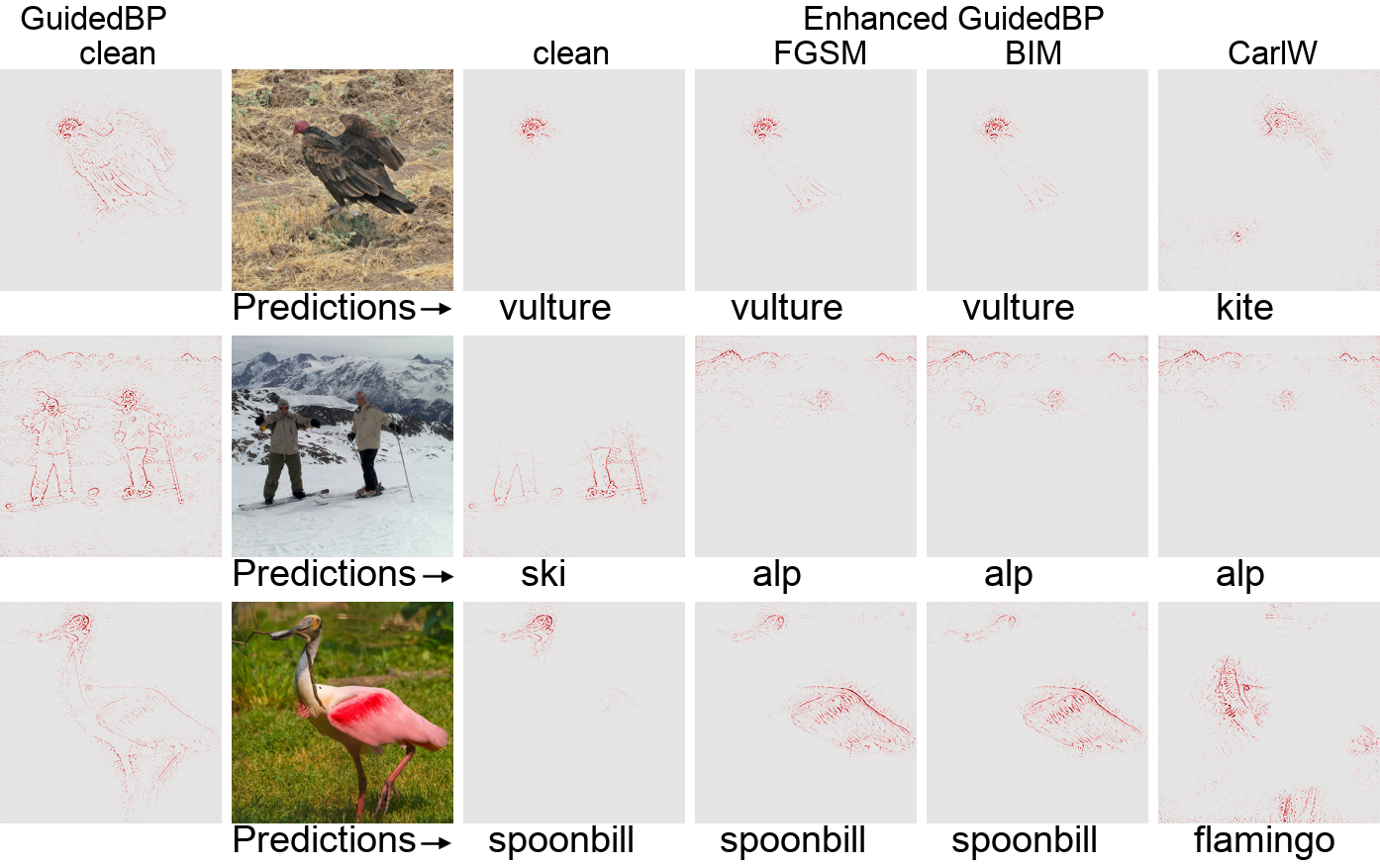}
   \caption{The figure shows SMs created by GuidedBP and Enhanced GuidedBP for clean images and adversary ones. The predictions under the map indicate the success or failure of adversary attacks.}
   \label{fig:fur_advdis}
   \vspace{-0.2cm}
\end{wrapfigure}

To further analyze the adversary-discriminativity of SMs created by enhanced GuidedBP, we categorize created adversary images into two categories: $Adv_f$ are the ones that fail to attack the neural network and $Adv_s$ are the ones that mislead the classification decisions successfully. For the clean images and the perturbed images in $Adv_f$, the created SMs should identify the class-discriminative parts. Contrarily, for the adversary images $Adv_s$, the parts identified in the SMs are irrelevant to the ground-truth label, which means the network focuses on the wrong parts of the adversary images when making decisions.

In Figure \ref{fig:fur_advdis}, the image in the first row contains a vulture. If the created adversary image fails to fool the neural network, the corresponding SM focuses on the head of the vulture (see 1st-3rd columns right of the image). If the attack is successful, the created SM for the misclassified class (i.e., kite) focuses on wings of the vulture. As a comparison, the GuidedBP always visualizes all the salient low-level features of all the images (e.g., the ski, the persons, and the alp in the image of the second row).

\section{Discussion and Conclusion}
\label{sec:disc}
\indent \indent \textbf{Why is enhanced GuidedBP better?} The pre-softmax scores (logits) are often taken as output scores to create SMs. The previous attribution methods show that the scores of different classes can be attributed to the same pixels. They explain where the scores themselves come from. Our approach explains where the difference between logits comes from, which is the exact reason why the network predicts a higher probability for a particular class, rather than another one. In the optimization of creating adversary images, the loss of the neural network is increased, which results in the change of the rank of logits. Our approach can find the evidence for the difference between the scores, i.e., the rank of logits. The change of the rank is the reason for misclassifications. That is why the enhanced GuidedBP can explain the classification decisions of adversary images better.

\textbf{The generalization and limitation of the enhancing method}
As shown in Sec. \ref{sec:dis_info}, the important factor to support the success of enhanced GuidedBP is that $\pmb{s}^k$ and $\pmb{s}^m$ have similar filtering effects. When generalizing the enhancing method to other methods, the effectiveness depends on the similarity of filtering effects. \cite{Ghorbani2017InterpretationON,Dombrowski2019ExplanationsCB} show that SMs can be manipulated due to vulnerability of DNNs. The limitation of the method is that we assume that the attack methods are not aware of our method.

\textbf{Conclusion} In this work, we identify the class-discriminative information in SMs created by GuidedBP and propose a simple way to enhance it. The proposed enhanced GuidedBP can explain classification decisions of adversary images better. In future work, we will investigate how to regularize the deep neural networks using the captured discriminative information so that the rank of logits is not easily changed by adversary perturbations.

\bibliography{neurips_2019}
\bibliographystyle{unsrt}

\end{document}